\documentclass[11pt,a4paper]{article}
\usepackage[hyperref]{acl2019}
\usepackage{times}
\usepackage{latexsym}
\usepackage{soul}
\usepackage{url}
 \usepackage{todonotes}
 \usepackage{amsmath}
\aclfinalcopy 

\newcommand{\transFHK}{\textit{transference500K}}
\newcommand{\transAll}{\textit{transferenceALL}}

\title{UDS--DFKI Submission to the WMT2019 Similar Language Translation Shared Task}

 \author{Santanu Pal\textsuperscript{1,3}, Marcos Zampieri\textsuperscript{2}, Josef van Genabith\textsuperscript{1,3}\\
 \textsuperscript{1}Department of Language Science and Technology, Saarland University, Germany\\
 \textsuperscript{2}Research Institute for Information and Language Processing, University of Wolverhampton, UK\\
  \textsuperscript{3}German Research Center for Artificial Intelligence (DFKI),\\ Saarland Informatics Campus, Germany \\
{\tt santanu.pal@uni-saarland.de}\\
  }

\date{}

\begin{document}
\maketitle
\begin{abstract}
In this paper we present the UDS-DFKI system submitted to the Similar Language Translation shared task at WMT 2019. The first edition of this shared task featured data from three pairs of similar languages: Czech and Polish, Hindi and Nepali, and Portuguese and Spanish. Participants could choose to participate in any of these three tracks and submit system outputs in any translation direction. We report the results obtained by our system in translating from Czech to Polish and comment on the impact of out-of-domain test data in the performance of our system. UDS-DFKI achieved competitive performance ranking second among ten teams in Czech to Polish translation. 
\end{abstract}

\section{Introduction} 

The shared tasks organized annually at WMT provide important benchmarks used in the MT community. Most of these shared tasks include English data, which contributes to make English the most resource-rich language in MT and NLP. In the most popular WMT shared task for example, the News task, MT systems have been trained to translate texts from and to English \cite{bojar2016findings,bojar2017findings}. 

This year, we have observed a shift on the dominant role that English on the WMT shared tasks. The News task featured for the first time two language pairs which did not include English: German-Czech and French-German. In addition to that,  the Similar Language Translation was organized for the first time at WMT 2019 with the purpose of evaluating the performance of MT systems on three pairs of similar languages from three different language families: Ibero-Romance, Indo-Aryan, and Slavic. 

The Similar Language Translation \cite{barrault-EtAl:2019:WMT} task provided participants with training, development, and testing data from the following language pairs: Spanish - Portuguese (Romance languages), Czech - Polish (Slavic languages), and Hindi - Nepali (Indo-Aryan languages). Participant could submit system outputs to any of the three language pairs in any direction. The shared task attracted a good number of participants and the performance of all entries was evaluated using popular MT automatic evaluation metrics, namely BLEU \cite{Papineni:2002:BMA:1073083.1073135} and TER \cite{Snover06astudy}.

In this paper we describe the UDS-DFKI system to the WMT 2019 Similar Language Translation task. The system achieved competitive performance and ranked second among ten entries in Czech to Polish translation in terms of BLEU score.

\section{Related Work} 

With the widespread use of MT technology and the commercial and academic success of NMT, there has been more interest in training systems to translate between languages other than English \cite{ruiz2017catalan}. One reason for this is the growing need of direct translation between pairs of similar languages, and to a lesser extent language varieties, without the use of English as a pivot language. The main challenge is to overcome the limitation of available parallel data taking advantage of the similarity between languages. Studies have been published on translating between similar languages (e.g. Catalan - Spanish \cite{ruiz2017catalan}) and language varieties such as European and Brazilian Portuguese \cite{fancellu2014standard,costa2018neural}. The study by \newcite{lakew2018neural} tackles both training MT systems to translate between European--Brazilian Portuguese and European--Canadian French, and two pairs of similar languages Croatian--Serbian and Indonesian--Malay.

Processing similar languages and language varieties has attracted attention not only in the MT community but in NLP in general. This is evidenced by a number of research papers published in the last few years and the recent iterations of the VarDial evaluation campaign which featured multiple shared tasks on topics such as dialect detection, morphosyntactic tagging, cross-lingual parsing, cross-lingual morphological analysis \cite{vardial2018report,vardial2019report}.

\section{Data}

We used the Czech--Polish dataset provided by the WMT 2019 Similar Language Translation task organizers for our experiments. The released parallel dataset consists of out-of-domain (or general-domain) data only and it differs substantially from the released development set which is part of a TED corpus. The parallel data includes Europarl v9, Wiki-titles v1, and JRC-Acquis. We combine all the released data and prepare a large out-domain dataset. 

\subsection{Pre-processing}

The out-domain data is noisy for our purposes,
so we apply methods for cleaning. We performed the following two steps: (i) we use the cleaning process described in \newcite{Pal:2015:WMT}, and (ii) we execute the Moses~\cite{Koehn:2007} corpus cleaning scripts with minimum and maximum number of tokens set to 1 and 100, respectively. After cleaning, we perform punctuation normalization, and then we use the Moses tokenizer to tokenize the out-domain corpus with `no-escape' option. Finally, we apply true-casing.

The cleaned version of the released data, i.e., the General corpus containing 1,394,319 sentences, is sorted based on the score in Equation \ref{eq:ce}.
Thereafter, We split the entire data (1,394,319) into two sets; we use the first 1,000 for validation and the remaining as training data. The released development set (Dev) is used as test data for our experiment. It should be noted noted that, we exclude 1,000 sentences from the General corpus which are scored as top (i.e., more in-domain like) during the data selection process.

We prepare two parallel training sets from the aforementioned training data: (i) \transFHK (presented next), collected 500,000 parallel data through data selection method \cite{Axelrod:2011}, which are very similar to the in-domain data (for our case the development set), and (ii) \transAll, utilizing all the released out-domain data sorted by Equation \ref{eq:ce}.

The \transFHK training set is prepared using in-domain (development set) bilingual cross-entropy difference for data selection as was described in~\newcite{Axelrod:2011}.
The difference in cross-entropy is computed based on two language models (LM): a domain-specific LM is estimated from the in-domain (containing 2050 sentences)  corpus ($lm_{i}$) and the out-domain LM ($lm_{o}$) is estimated from the eScape corpus. We rank the eScape corpus by assigning a score to each of the individual sentences which is the sum of the three cross-entropy ($H$) differences. For a $j^{th}$ sentence pair ${src}_j$--${trg}_j$, the score is calculated based on Equation~\ref{eq:ce}.  

\begin{multline}
score = |H_{src}({src}_j, lm_{i}) - H_{src}(src_j, lm_{o})| \\
+ |H_{trg}(trg_j, lm_{i}) - H_{trg}(trg_j, lm_{o})|
\label{eq:ce}
\end{multline}

\vspace{2mm}

\section{System Architecture - The Transference Model}

Our \textit{transference} model extends the original transformer model to multi-encoder based transformer architecture. The \textit{transformer} architecture~\cite{Vaswani:NIPS2017} is built solely upon such attention mechanisms completely replacing recurrence and convolutions. The transformer uses positional encoding to encode the input and output sequences, and computes both self- and cross-attention through so-called multi-head attentions, which are facilitated by parallelization. We use multi-head attention to jointly attend to information at different positions from different representation subspaces. 

The first encoder ($enc_1$) of our model encodes word form information of the source ($f_w$), and a second sub-encoder ($enc_2$) to encode sub-word (byte-pair-encoding) information of the source ($f_s$). 
Additionally, a second encoder ($enc_{src \rightarrow mt}$) which takes the encoded representation from the $enc_1$, combines this with the self-attention-based encoding of $f_s$ ($enc_2$), and prepares a representation for the decoder ($dec_{e}$) via cross-attention. Our second encoder ($enc_{1 \rightarrow 2}$) can be viewed as a transformer based NMT's decoding block, however, without masking. The intuition behind our architecture is to generate better representations via both self- and cross-attention and to further facilitate the learning capacity of the feed-forward layer in the decoder block. In our transference model, one self-attended encoder for $f_w$, $\mathbf{f_w}$ = $(w_1, w_2, \ldots, w_k)$, returns a sequence of continuous representations, $enc_{2}$, and a second self-attended sub-encoder for $f_s$, $\mathbf{f_s}$ = $(s_1, s_2, \ldots, s_l)$, returns another sequence of continuous representations, $enc_{2}$. Self-attention at this point provides the advantage of aggregating information from all of the words, including $f_w$ and $f_s$, and successively generates a new representation per word informed by the entire $f_w$ and $f_s$ context. The internal $enc_{2}$ representation performs cross-attention over $enc_{1}$ and prepares a final representation ($enc_{1 \rightarrow 2}$) for the decoder ($dec_{e}$). The decoder generates the $e$ output in sequence, $\mathbf{e}$ = $(e_1, e_2, \ldots, e_n)$, one word at a time from left to right by attending to previously generated words as well as the final representations ($enc_{1 \rightarrow 2}$) generated by the encoder.

We use the scale-dot attention mechanism (like \newcite{Vaswani:NIPS2017}) for both self- and cross-attention, 
as defined in Equation \ref{eq:1}, where $Q$, $K$ and $V$ are query, key and value, respectively, and $d_k$ is the dimension of $K$.
\begin{equation}
    attention(Q,K,V) = softmax(\frac{QK^T}{\sqrt{d_k}})V
    \label{eq:1}
\end{equation}

\noindent The multi-head attention mechanism in the transformer network maps the Q, K, and V matrices by using different linear projections. Then $h$ parallel heads are employed to focus on different parts in V.
The $i^{th}$ multi-head attention is denoted by $head_i$ in Equation~\ref{eq:2}. $head_i$ is linearly learned by three projection parameter matrices: $W_i^Q,W_i^K \in R^{d_{model} \times d_k}$, $W_i^V \in R^{d_{model} \times d_v}$; where $d_k = d_v = d_{model}/h$, and $d_{model}$ is the number of hidden units of our network.
\begin{equation}
    head_i = attention(QW_i^Q, KW_i^K, VW_i^V) 
    \label{eq:2}
\end{equation}

\noindent Finally, all the vectors produced by parallel heads are linearly projected using concatenation and form a single vector, called a multi-head attention ($M_{att}$) (cf.~Equation \ref{eq:3}). Here the dimension of the learned weight matrix $W^O$ is $R^{d_{model} \times d_{model}}$.
\begin{equation}
    M_{att}(Q,K,V) = Concat_{i:1}^n(head_i)W^O 
    \label{eq:3}
\end{equation}

\vspace{2mm}

\section{Experiments}

We explore our \textit{transference} model --a two-encoder based transformer architecture, in CS-PL similar language translation task. 

\subsection{Experiment Setup}
\label{sec:exp_setup}

For \transAll, we initially train on the complete out-of-domain dataset (General). The General data is sorted based on their in-domain similarities as described in Equation \ref{eq:ce}. 

\transAll models are then fine-tuned towards the 500K (in-domain-like) data. Finally, we perform checkpoint averaging using the 8 best checkpoints.
We report the results on the provided development set, which we use as a test set before the submission. Additionally we also report the official test set result.

To handle out-of-vocabulary words and to reduce the vocabulary
size, instead of considering words, we consider subword units~\cite{Sennrich:2016ACL} by using byte-pair encoding (BPE). In the preprocessing step, instead of learning an explicit mapping between BPEs in the
Czech (CS) and Polish (PL), we define BPE tokens by jointly processing all parallel data. Thus, CS and PL derive a single BPE vocabulary.
Since CS and PL belong to the similar language, they naturally share a good fraction of BPE tokens, which reduces the vocabulary size. 

We pass word level information on the first encoder and the BPE information to the second one. On the decoder side of the transference model we pass only BPE text.
 
We evaluate our approach with development data which is used as test case before submission. We use BLEU \cite{Papineni:2002:BMA:1073083.1073135} and TER \cite{Snover06astudy}.  
 
\subsection{Hyper-parameter Setup}

We follow a similar hyper-parameter setup for all reported systems.
All encoders, and the decoder, are composed of a stack of $N_{fw} = N_{fs} = N_{es} = 6$ identical layers followed by layer normalization. 
Each layer again consists of two sub-layers and a residual connection~\cite{He:2015:Deep} around each of the two sub-layers. We apply dropout~\cite{Srivastava:2014:DSW} to the output of each sub-layer, before it is added to the sub-layer input and normalized. 
Furthermore, dropout is applied to the sums of the word embeddings and the corresponding positional encodings in both encoders as well as the decoder stacks.

We set all dropout values in the network to 0.1. 
During training, we employ label smoothing with value $\epsilon_{ls}$ = 0.1. 
The output dimension produced by all sub-layers and embedding layers is $d_{model} = 512$. Each encoder and decoder layer contains a fully connected feed-forward network ($FFN$) having dimensionality of $d_{model} = 512$ for the input and output and dimensionality of $d_{ff} = 2048$ for the inner layers. For the scaled dot-product attention, the input consists of queries and keys of dimension $d_k$, and values of dimension $d_v$. As multi-head attention parameters, we employ $h = 8$ for parallel attention layers, or heads. For each of these we use a dimensionality of $d_k = d_v = d_{model}/h = 64$. 
For optimization, we use the Adam optimizer~\cite{Kingma:2014method} with $\beta_1 = 0.9$, $\beta_2 = 0.98$ and $\epsilon = 10^{-9}$. 

The learning rate is varied throughout the training process, and increasing for the first training steps $warmup_{steps} = 8000$ and afterwards decreasing as described in~\cite{Vaswani:NIPS2017}. All remaining hyper-parameters are set analogously to those of the transformer's \textit{base} model.
At training time, the batch size is set to 25K tokens, with a maximum sentence length of 256 subwords, and a vocabulary size of 28K. After each epoch, the training data is shuffled. 
After finishing training, we save the 8 best checkpoints which are written at each epoch. Finally, we use a single model obtained by averaging the last 8 checkpoints. 
During decoding, we perform beam search with a beam size of 4. 
We use shared embeddings between CS and PL in all our experiments.

\section{Results}

We present the results obtained by our system in Table \ref{result:dev-test}. 

\begin{table}[!ht]
\centering
\begin{tabular}{lccc}
\hline
\bf tested on   & model     & \bf BLEU & \bf TER  \\ 
\hline
Dev set & Generic & 12.2 & 75.8 \\
Dev set & Fine-tuned* & \textbf{25.1} & \textbf{58.9} \\
Test set & Generic & 7.1 & 89.3 \\
Test set & Fine-Tuned* & \textbf{7.6} & \textbf{87.0} \\
\hline
\end{tabular}
\caption{Results for CS--PL Translation; * averaging 8 best checkpoints.}
\label{result:dev-test}
\end{table}

\noindent Our fine-tuned system on development set provides significant performance improvement over the generic model. We found +12.9 absolute BLEU points improvement over the generic model. Similar improvement is also observed in terms of TER (-16.9 absolute). It is to be noted that our generic model is trained solely on the clean version of training data.

Before submission, we performed punctuation normalization, unicode normalization, and detokenization for the run.

In Table \ref{result:rank} we present the ranking of the competition provided by the shared task organizers. Ten entries were submitted by five teams and are ordered by BLEU score. TER is reported for all submissions which achieved BLEU score greater than 5.0. The type column specifies the type of system, whether it is a Primary (P) or Constrastive (C) entry.

\begin{table}[!ht]
\centering
\begin{tabular}{lccc}
\hline
\bf Team        & \bf Type & \bf BLEU & \bf TER  \\ 
\hline
UPC-TALP	& P	& 7.9	& 85.9 \\
\textbf{UDS-DFKI}	& P	& \textbf{7.6}	& \textbf{87.0} \\
Uhelsinki	& P & 7.1	& 87.4 \\
Uhelsinki	& C	& 7.0	& 87.3 \\ 
Incomslav	& C	& 5.9	& 88.4 \\
Uhelsinki	& C	& 5.9	& 88.4 \\
Incomslav	& P	& 3.2	& - \\
Incomslav	& C	& 3.1	& - \\
UBC-NLP	& C	& 2.3	& - \\
UBC-NLP	& P	& 2.2	& - \\

\hline
\end{tabular}
\caption{Rank table for Czech to Polish Translation}
\label{result:rank}
\end{table}

\noindent Our system was ranked second in the competition only 0.3 BLEU points behind the winning team UPC-TALP. The relative low BLEU and high TER scores obtained by all teams are due to out-of-domain data provided in the competition which made the task equally challenging to all participants. 

\section{Conclusion}

This paper presented the UDS-DFKI system submitted to the Similar Language Translation shared task at WMT 2019. We presented the results obtained by our system in translating from Czech to Polish. Our system achieved competitive performance ranking second among ten teams in the competition in terms of BLEU score. The fact that out-of-domain data was provided by the organizers resulted in a challenging but interesting scenario for all participants. 

In future work, we would like to investigate how effective is the proposed hypothesis (i.e., word-BPE level information) in similar language translation. Furthermore, we would like to explore the similarity between these two languages (and the other two language pairs in the competition) in more detail by training models that can best capture morphological differences between them. During such competitions, this is not always possible due to time constraints.

\section*{Acknowledgments}

This research was funded in part by the German research foundation (DFG) under grant number GE 2819/2-1 (project MMPE) and the German Federal Ministry of Education and Research (BMBF) under funding code 01IW17001 (project Deeplee). The responsibility for this publication lies with the authors. We would like to thank the anonymous WMT reviewers for their valuable input, and the organizers of the shared task.

\bibliography{acl2019}
\bibliographystyle{acl_natbib}

\end{document}